\documentclass{article}
\usepackage[table]{xcolor}
    \PassOptionsToPackage{numbers, compress}{natbib}
\usepackage[numbers]{natbib}
 \usepackage[preprint]{neurips_2025}

\usepackage[utf8]{inputenc} 
\usepackage[T1]{fontenc}    
\usepackage{hyperref}       
\usepackage{url}            
\usepackage{booktabs}       
\usepackage{amsfonts}       
\usepackage{nicefrac}       
\usepackage{microtype}      
\usepackage{xcolor}         

\usepackage{graphicx}
\usepackage{subcaption}
\usepackage{booktabs}
\usepackage{xspace}
\usepackage{natbib}
\usepackage{makecell}
\usepackage{multirow}
\usepackage{algorithm}
\usepackage{algorithmic}

\usepackage{tabularx}
\usepackage{array}
\usepackage{booktabs}

\newcommand{\red}[1]{\textcolor{red}{#1}}
\newcommand{\blue}[1]{\textcolor{blue}{#1}}

\definecolor{rule}{HTML}{F27970}
\definecolor{model}{HTML}{BB9727}
\definecolor{answer}{HTML}{54B345}
\definecolor{process}{HTML}{05B9E2}
\definecolor{backred}{RGB}{255, 190, 190}
\definecolor{backblue}{RGB}{210, 230, 250}

\newcommand{\best}[1]{\cellcolor{backred}\textbf{#1}}
\newcommand{\high}[1]{\cellcolor{backblue}\textbf{#1}}

\usepackage{pifont}

\newcommand{\greenyes}{\textcolor{green}{\ding{51}}}
\newcommand{\redno}{\textcolor{red}{\ding{55}}}

\usepackage{enumitem}

\usepackage{tcolorbox}
\tcbuselibrary{most}
\definecolor{lightgreen}{RGB}{200,255,200}
\definecolor{lightpink}{rgb}{1.0, 0.85, 0.9} 
\definecolor{lightblue}{rgb}{0.529, 0.808, 0.922} 
\definecolor{lightgray}{gray}{0.85}

\usepackage{wrapfig}

\newcommand{\dataset}{\textsc{MMSpec}\xspace}
\newcommand{\method}{\textsc{ViSkip}\xspace}

\newtcolorbox[auto counter]{takeaway}[1][]{
    enhanced,
    breakable,
    colframe=lightblue,
    colback=lightblue!30!white,
    sharp corners,
    boxsep=0pt,
    left=5pt,
    right=5pt,
    top=6pt,
    bottom=6pt,
    boxrule=0pt,
    leftrule=4pt,
    before upper={\textbf{Takeaway \thetcbcounter: } },
    #1
}

\title{\begin{minipage}{.08\textwidth}
\centering
\vspace{-4pt}
\includegraphics[width=\linewidth]{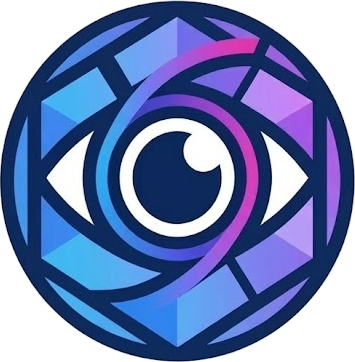} 
\end{minipage} \dataset: Benchmarking Speculative Decoding for Vision-Language Models}

\author{Hui Shen$^{1,*}$, Xin Wang$^{2,*}$, Ping Zhang$^{2,*}$, Yunta Hsieh$^{1}$,\vspace{0.1cm}\\ \textbf{Qi Han$^3$, Zhongwei Wan$^2$, Ziheng Zhang$^2$, Jingxuan Zhang$^{4}$, Jing Xiong$^{5}$,}\vspace{0.1cm}\\ \textbf{Ziyuan Liu$^{6}$, Yifan Zhang$^{1}$, Hangrui Cao$^{7}$, Chenyang Zhao$^{8}$, Mi Zhang$^{2}$}\vspace{0.3cm}\\ 
  $^1$University of Michigan, 
  $^2$The Ohio State University, 
  $^3$Independent,
  $^4$Indiana University,\\
  $^5$The University of Hong Kong
  $^6$Peking University,
  $^7$Carnegie Mellon University,
  $^8$LMSYS Org
}

\begin{document}

\maketitle

\begin{abstract}
Vision-language models (VLMs) achieve strong performance on multimodal tasks but suffer from high inference latency due to large model sizes and long multimodal contexts. Speculative decoding has recently emerged as an effective acceleration technique, yet its behavior in VLMs remains insufficiently understood.  We introduce \dataset, the first benchmark for evaluating speculative decoding in vision-language models. \dataset contains 600 multimodal samples across six task categories and integrates ten representative speculative decoding algorithms under a unified evaluation framework. Our study reveals three key findings: (1) methods designed for text-only LLMs degrade in multimodal scenarios, (2) vision awareness becomes increasingly important at larger batch sizes, and (3) throughput speedup alone does not reliably reflect latency performance. Motivated by these findings, we propose \method, a plug-and-play speculative decoding method that dynamically adapts speculation to vision tokens and achieves state-of-the-art performance. More details are available on our project page: \href{https://killthefullmoon.github.io/projects/MMSpec/index.html}{mmspec-bench.github.io}.
\end{abstract}

\section{Introduction}

Vision-Language Models (VLMs) have rapidly advanced the state of the art in multimodal reasoning, visual question answering, and multimodal content generation. By integrating visual perception with language understanding, modern VLMs enable a wide range of applications, including multimodal assistants, document understanding, and embodied intelligence. However, these models suffer from substantial inference latency due to their large model sizes, long multimodal contexts, and the inherently sequential nature of autoregressive decoding. This latency bottleneck poses a significant challenge for deploying VLMs in real-world interactive systems.

Speculative decoding has recently emerged as one of the most effective techniques for accelerating autoregressive generation in large language models. By generating draft tokens using a lightweight approximation and verifying them with the target model in parallel, speculative decoding can significantly improve decoding efficiency while preserving the exact output distribution. Motivated by its success in text-only LLMs, a growing number of works have begun extending speculative decoding to vision-language models.

Despite these advances, speculative decoding for vision-language models remains insufficiently studied, primarily due to the lack of a comprehensive and standardized evaluation benchmark. Existing evaluations are almost exclusively conducted on text-only datasets, which fail to capture key characteristics of multimodal generation, such as cross-modal dependencies, visually grounded reasoning, and heterogeneous multimodal context structures. Moreover, prior works evaluate their methods using different datasets, models, and experimental setups, making direct comparisons across approaches difficult. As a result, researchers and practitioners lack clear guidance on which speculative decoding methods are most effective in vision-language scenarios.

\begin{wrapfigure}{r}{0.5\linewidth}
    \centering
    \includegraphics[width=0.9\linewidth]{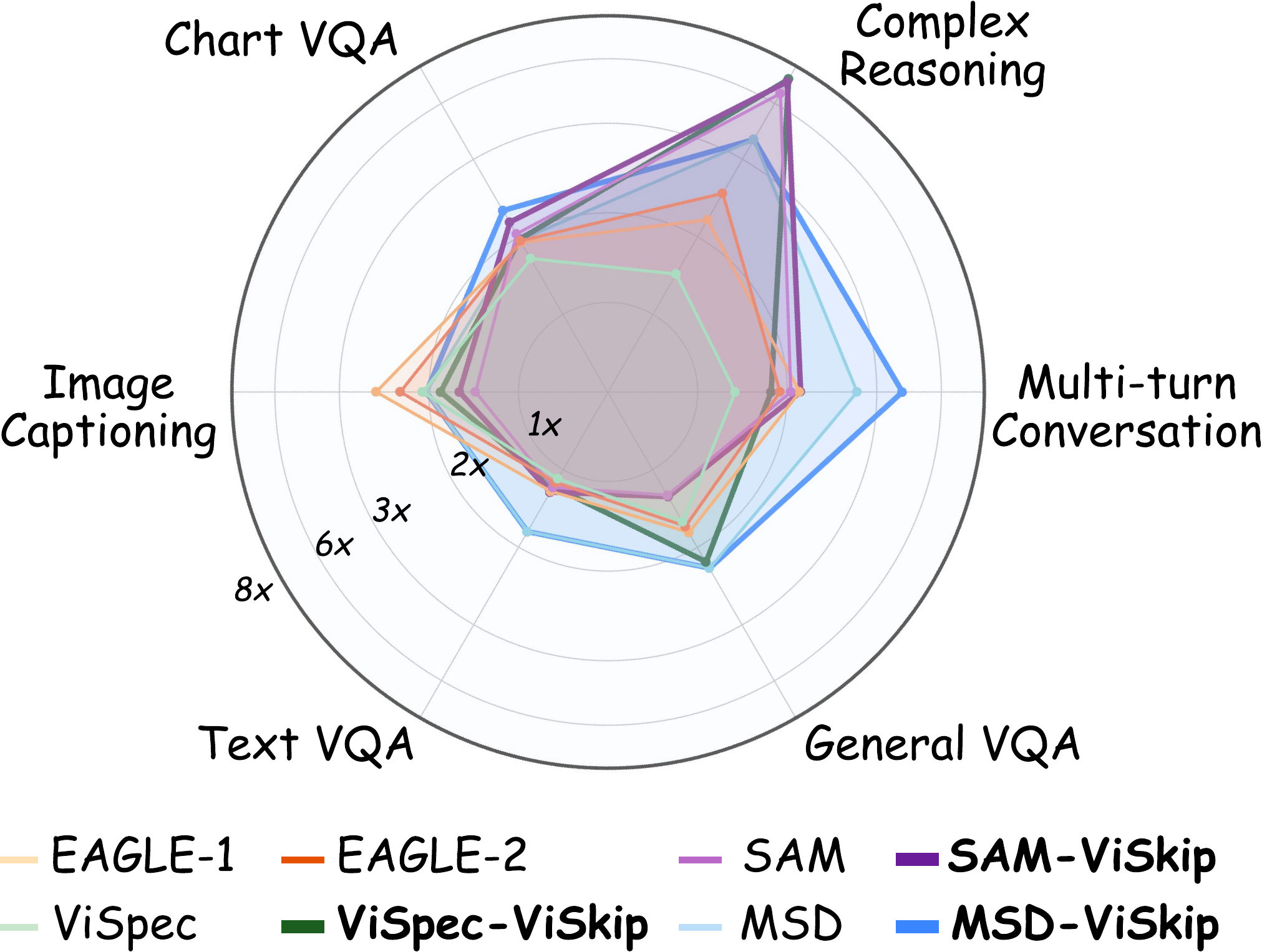}
    \caption{Performance comparison of speculative decoding methods with Qwen2.5-VL-7B on \dataset benchmark.}
    \label{fig:radar}
\end{wrapfigure}

To address this gap, we present \dataset, the first comprehensive benchmark and unified evaluation platform for speculative decoding in vision-language models. \dataset consists of 600 carefully curated multimodal samples spanning six representative task categories, covering diverse visual complexity, reasoning requirements, and output lengths. In addition, we implement a unified evaluation platform that integrates 10 representative speculative decoding algorithms—including draft-model-based methods, history-based methods, and multimodal-aware methods—under a consistent experimental framework, enabling fair and reproducible comparisons.

Using \dataset, we conduct the first systematic empirical study of speculative decoding for VLMs and obtain three key findings. 
First, speculative decoding methods designed for text-only LLMs often experience noticeable performance degradation in multimodal scenarios due to cross-modal dependencies. 
Second, vision awareness plays an increasingly important role as batch size grows; methods that fail to incorporate visual information suffer from reduced draft accuracy and diminished speedups, while vision-aware approaches remain more stable. 
Third, throughput speedup alone is insufficient to evaluate speculative decoding performance, as methods with higher speedups do not always achieve better latency behavior. These observations highlight the importance of evaluating speculative decoding from both throughput and latency perspectives.

Based on these findings, we develop \method, a plug-and-play speculative decoding method that dynamically adapts speculation to vision tokens. As shown in Figure~\ref{fig:radar}, \method achieves state-of-the-art performance compared with baselines.

Our contributions are summarized as follows:

\begin{itemize}

\item We introduce \dataset, the first comprehensive benchmark for evaluating speculative decoding in vision-language models, covering diverse multimodal workloads and generation characteristics.

\item We build a unified evaluation platform integrating ten representative speculative decoding algorithms under a consistent and reproducible framework.

\item We conduct the first systematic empirical study of speculative decoding in vision-language settings, revealing key limitations and unique behaviors of existing methods.

\item Based on these findings, we develop \method, a plug-and-play speculative decoding method that achieves state-of-the-art performance on VLM inference.

\end{itemize}
\section{Related Work}

\subsection{Efficient Vision-Language Models}

Vision-language models (VLMs) have achieved strong performance across various multimodal tasks, such as visual question answering, image captioning, and multimodal reasoning. However, their inference efficiency remains a major challenge due to large model sizes and the additional computational cost introduced by visual tokens. Prior work improves VLM efficiency through both architectural and system-level optimizations, including visual token compression, efficient attention mechanisms, and optimized inference systems. More recently, speculative decoding has been extended to VLMs to accelerate autoregressive generation. Multimodal-aware designs, such as MSD-style methods~\cite{lin2025speculativedecodingreimaginedmultimodal} and ViSpec~\cite{kang2025vispecacceleratingvisionlanguagemodels}, adapt drafting strategies to the vision-language structure and demonstrate improved efficiency compared with directly applying LLM-oriented speculative decoding methods.

\subsection{Speculative Decoding}

Speculative decoding was formalized as a draft-then-verify generation paradigm that accelerates autoregressive inference while preserving the exact output distribution under correct verification~\cite{leviathan2023fastinferencetransformersspeculative}. In this framework, a lightweight draft model proposes multiple candidate tokens which are subsequently verified by the full model, allowing several decoding steps to be executed in parallel.

Follow-up works mainly differ in how draft candidates are generated and verified. Medusa~\cite{cai2024medusasimplellminference} proposes a multi-head drafting architecture that attaches auxiliary prediction heads to the base model, enabling multiple future tokens to be predicted simultaneously. EAGLE and EAGLE-2~\cite{li2025eaglespeculativesamplingrequires,li2024eagle2fasterinferencelanguage} introduce feature-level drafting with uncertainty-aware mechanisms and dynamic draft trees, improving acceptance rates and inference efficiency. Lookahead decoding~\cite{fu2024breaksequentialdependencyllm} explores a parallel decoding paradigm that reduces sequential dependency without requiring a separate draft model. 

Another line of work focuses on training-free speculative methods. Prompt Lookup Decoding reuses repeated n-grams from the prompt or context to generate draft tokens without additional training or auxiliary models~\cite{mangrulkar2023promptlookup}. These approaches exploit redundancy in the generation process to obtain speedups while maintaining exact decoding.

More recently, speculative decoding has been extended beyond pure language models to multimodal settings. However, applying standard speculative decoding to vision-language models introduces new challenges due to cross-modal dependencies between visual tokens and generated text. As a result, naive drafting strategies often suffer from higher rejection rates when the generation strongly depends on visual grounding. This observation has motivated recent multimodal-aware speculative decoding methods that adapt the drafting process to the structure of VLMs~\cite{lin2025speculativedecodingreimaginedmultimodal,kang2025vispecacceleratingvisionlanguagemodels}.

Despite the advancements, existing evaluations~\citep{xia2024unlocking} of speculative decoding are almost exclusively conducted on text-only benchmarks. Such settings fail to capture the unique characteristics of multimodal generation, including cross-modal dependencies, visually grounded reasoning, and heterogeneous multimodal context structures. 

Moreover, prior studies typically evaluate their methods using different datasets, models, and experimental setups, which makes direct comparisons across approaches difficult. As a result, the community currently lacks a unified evaluation framework that systematically assesses speculative decoding methods in vision-language scenarios, leaving researchers and practitioners without clear guidance on which approaches are most effective for multimodal generation.

\section{The \dataset Benchmark}
\begin{figure}[t]
    \centering
    \includegraphics[width=1\linewidth]{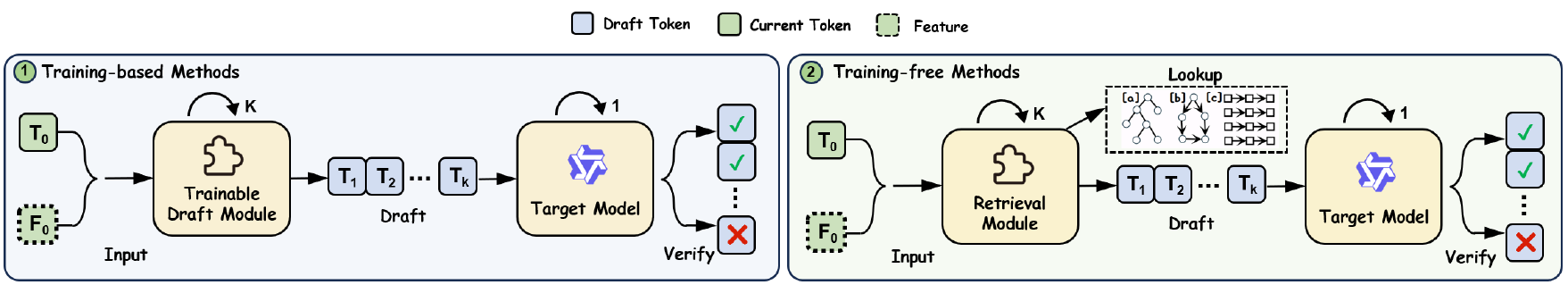}
    \caption{Overview of speculative decoding algorithms evaluated in MMSpec framework.}
    \label{fig:framework}
\end{figure}

With the rapid progress in multimodal speculative decoding, there is a growing need for rigorous comparisons of leading methods. However, existing approaches are evaluated on inconsistent benchmarks, devices, and environments, making fair comparisions impractical. To address this gap, we introduce \dataset, a comprehensive benchmark for multimodal speculative inference that spans diverse input modalities and real-world application settings. Built on MMSpec, we conduct a systematic third-party evaluation of representative open-source approaches, using a unified device, identical software environment, and standardized evaluation protocol to ensure reproducible and fair comparisons. 
This section outlines the two core components of our experimental setup: the evaluation datasets and the vision-language speculative decoding algorithms compared. The overall framework is illustrated in Figure~\ref{fig:framework}.

\begin{figure}
    \centering
    \includegraphics[width=1\linewidth]{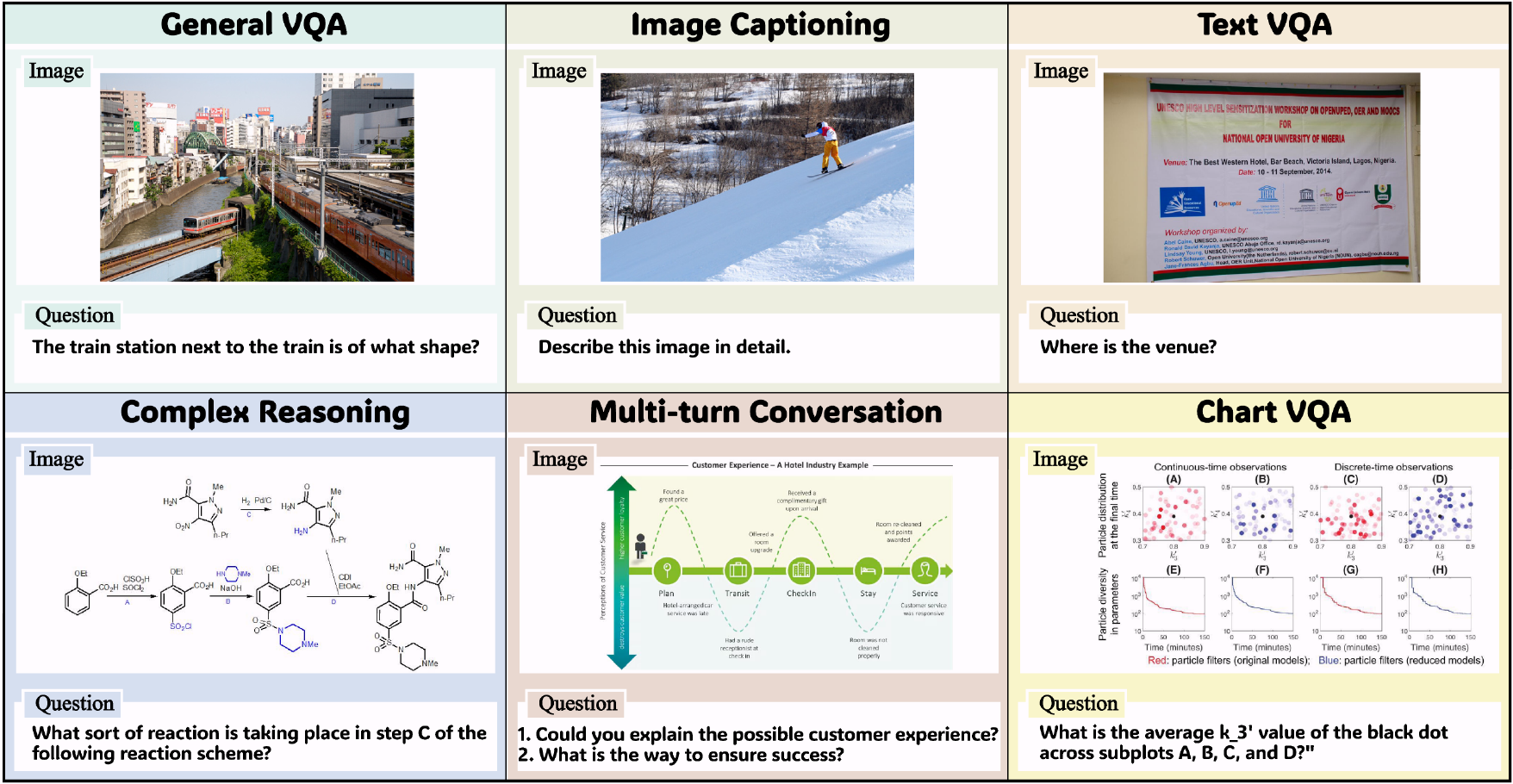}
    \caption{Sampled data from \dataset}
    \label{fig:data_sample}
\end{figure}

\subsection{Dataset Construction}
To assess vision-language speculative decoding methods across various scenarios, we enforce four principles: (i) workload diversity, (ii) balanced topic coverage, (iii) explicit multi-turn support, and (iv) method-agnostic measurement. \dataset covering six distinct subtasks: \textit{General VQA}, \textit{Text VQA}, \textit{Image Captioning}, \textit{Chart VQA}, \textit{Complex Reasoning}, \textit{Multi-turn Conversation} with representative examples illustrated in Figure~\ref{fig:data_sample}.
We composed the dataset by selecting 100 instances from the following datasets:
\begin{itemize}
    [label=$\bullet$, leftmargin=*, itemsep=0pt, parsep=2pt]
    \item \textbf{GQA}~\citep{hudson2019gqanewdatasetrealworld}: GQA is a large-scale benchmark for real-world visual reasoning and compositional question answering, built from scene-graph-grounded questions that emphasize object attributes, relations, and multi-step reasoning. 

    \item \textbf{TextVQA}~\citep{singh2019vqamodelsread}: TextVQA evaluates whether models can read and reason over text embedded in natural scenes. 

    \item \textbf{COCO}~\citep{lin2015microsoftcococommonobjects}: MS COCO is a foundational benchmark of everyday scenes in natural context and has become one of the standard sources for image captioning evaluation.

    \item \textbf{CharXiv}~\citep{wang2024charxivchartinggapsrealistic}: CharXiv is a realistic chart-understanding benchmark built from charts extracted from arXiv papers, with both descriptive and reasoning questions over diverse scientific figures. 

    \item \textbf{MMMU-Pro}~\citep{yue2025mmmuprorobustmultidisciplinemultimodal}: MMMU-Pro is a more robust multi-discipline multimodal reasoning benchmark that filters out text-only shortcuts, strengthens answer options, and introduces harder vision-only settings. 

    \item \textbf{ConvBench}~\citep{liu2024convbenchmultiturnconversationevaluation}: ConvBench is a multi-turn conversation benchmark for vision-language models, organized around a hierarchy of perception, reasoning, and creativity to diagnose errors across turns. 

    \item \textbf{MM-MT-Bench}~\citep{mmmtbench_dataset}: MM-MT-Bench is an open-ended multi-turn multimodal benchmark for evaluating instruction following in practical conversational scenarios. 
\end{itemize}

This benchmark serves as a challenging test for our core experiments. Table~\ref{tab:dataset_composition} summarizes the data distribution.

\begin{table}[t]
\centering
\caption{\dataset composition, avg. output length computed from Qwen2.5-VL-7B.}
\label{tab:dataset_composition}
\setlength{\tabcolsep}{4pt}
\renewcommand{\arraystretch}{1.05}
\small
\begin{tabularx}{\linewidth}{l c X c}
\toprule
Topic & Samples & Data source(s) & Avg. output length (tokens) \\
\midrule
General VQA & 100 & GQA~\cite{hudson2019gqanewdatasetrealworld} & 46.98 \\
Text VQA & 100 & TextVQA~\cite{singh2019vqamodelsread} & 63.15 \\
Image Captioning & 100 & COCO~\cite{lin2015microsoftcococommonobjects} & 191.90 \\
Chart VQA & 100 & CharXiv~\cite{wang2024charxivchartinggapsrealistic} & 68.56 \\
Complex Reasoning & 100 & MMMU-Pro~\cite{yue2025mmmuprorobustmultidisciplinemultimodal} & 285.60 \\
Multi-turn Conversation & 100 &
ConvBench~\cite{liu2024convbenchmultiturnconversationevaluation} MMMTBench~\cite{mmmtbench_dataset}
& 747.65 \\
\midrule
Total & 600 & 7 sources  & 233.97 \\
\bottomrule
\end{tabularx}
\end{table}

\begin{table*}[htbp]
    \centering
    \caption{Summary of different speculative decoding methods. The table outlines each method's key idea, category (\textit{training-based}, \textit{Training-free}, or \textit{vision-aware}), speculation structure (\textit{Linear} or \textit{Tree}), ability to reuse repetitive content during generation, and whether the method is \textit{vision-aware} or \textit{vision-agnostic}.}
    \label{tab:speculative_methods_summary}
    \resizebox{1.0\textwidth}{!}{%
        \begin{tabular}{l p{6cm} c c c c}
            \toprule
            \multicolumn{1}{l}{\multirow{2}{*}{\textbf{Methods}}} 
            & \multirow{2}{*}{\textbf{Key Idea}} 
            & \multicolumn{2}{c}{\textbf{Drafting}} 
            & \multirow{2}{*}{\textbf{Vision Awareness}} 
            & \multirow{2}{*}{\textbf{Category}} \\
            \cmidrule(lr){3-4}
            & 
            & Structure & Reuse 
            &  &  \\
            \midrule

            \makecell[l]{ViSpec~\citep{kang2025vispecacceleratingvisionlanguagemodels}} & 
            Vision-token compression for efficient multimodal drafting. & 
            Linear & 
            \redno &
            Vision-aware &
            Training-based\\
            \addlinespace[1.5pt]

            \makecell[l]{MSD~\citep{lin2025speculativedecodingreimaginedmultimodal}} & 
            Train a multimodal draft with staged VLM training. & 
            Linear & 
            \redno &
            Vision-aware &
            Training-based\\

            \makecell[l]{EAGLE-1~\citep{DBLP:conf/icml/LiW0024}} & 
            Feature-level drafting from target hidden states. & 
            Linear & 
            \redno &
            Vision-agnostic &
            Training-based\\
            \addlinespace[1.5pt]

            \makecell[l]{EAGLE-2~\citep{DBLP:conf/emnlp/LiW0024}} & 
            Context-adaptive draft tree for higher acceptance. & 
            Tree & 
            \redno &
            Vision-agnostic &
            Training-based\\
            \addlinespace[1.5pt]

            \makecell[l]{EAGLE-3~\citep{DBLP:journals/corr/abs-2503-01840}} & 
            Token-level drafting with multi-layer feature fusion. & 
            Tree & 
            \redno &
            Vision-agnostic &
            Training-based\\
            \addlinespace[1.5pt]

            \makecell[l]{Medusa~\citep{DBLP:conf/icml/CaiLGPLCD24}} & 
            Multi-head tree proposals from one forward pass. & 
            Tree & 
            \redno &
            Vision-agnostic &
            Training-based\\
            \addlinespace[1.5pt]

            \makecell[l]{SAM-Decoding~\citep{DBLP:conf/acl/HuWZZLCZ25}} & 
            Suffix-automaton longest-match continuation drafting. & 
            Linear & 
            \greenyes &
            Vision-agnostic &
            Training-free\\
            \addlinespace[1.5pt]

            \makecell[l]{Recycling~\citep{DBLP:conf/acl/LuoWZZZ0X25}} & 
            Reuse discarded candidates as draft tree nodes. & 
            Tree & 
            \greenyes &
            Vision-agnostic &
            Training-free\\
            \addlinespace[1.5pt]

            \makecell[l]{PLD~\citep{saxena2023prompt}} & 
            Prompt n-gram lookup as drafts (no draft model). & 
            Linear & 
            \greenyes &
            Vision-agnostic &
            Training-free\\
            \addlinespace[1.5pt]

            \makecell[l]{Lookahead~\citep{DBLP:conf/kdd/ZhaoXLZG24}} & 
            Trie-based retrieval of multi-token continuations + tree verification/accept. & 
            Tree & 
            \greenyes &
            Vision-agnostic &
            Training-free\\
            \addlinespace[1.5pt]

            \bottomrule
        \end{tabular}%
    }
\end{table*}

\subsection{Speculative Decoding Algorithms}
Following the design from~\citep{xia2024unlocking}, we benchmark ten lossless speculative decoding algorithms that can be broadly categorized into two main groups: \textit{training-based methods} and \textit{training-free methods}. These categories reflect different strategies for generating draft tokens during speculative decoding. We summarize the evaluated methods in Table~\ref{tab:speculative_methods_summary} and provide detailed descriptions below.

\subsection{Training-based Methods} Training-based speculative decoding methods improve drafting quality by introducing additional learnable components or specialized training strategies. These methods typically modify the architecture of the base model or train an auxiliary model to generate higher-quality draft tokens, thereby increasing the acceptance rate during verification and improving overall decoding throughput.

\begin{itemize}[label=$\bullet$, leftmargin=*, itemsep=0pt, parsep=2pt]

\item \textbf{\textit{EAGLE-1/2/3}}~\citep{DBLP:conf/icml/LiW0024,DBLP:conf/emnlp/LiW0024,DBLP:journals/corr/abs-2503-01840}:  
EAGLE-1 enables speculative decoding without requiring a separate draft model by generating draft tokens through feature-level auto-regressive prediction. Instead of predicting tokens directly, EAGLE predicts intermediate hidden features and maps them to candidate tokens, which are then verified losslessly by the target model. EAGLE-2 further improves efficiency by introducing a context-aware dynamic draft tree that adapts the number of draft branches according to decoding uncertainty. EAGLE-3 addresses the feature-prediction bottleneck by moving toward more direct token-level drafting with multi-layer feature fusion, significantly improving scalability and throughput.

\item \textbf{\textit{Medusa}}~\citep{DBLP:conf/icml/CaiLGPLCD24}:  
Medusa augments the base language model with multiple light-weight prediction heads, each responsible for predicting tokens at different future positions. These heads collectively generate a small tree of candidate continuations in a single forward pass. The base model then verifies the candidate tree using a tree-aware verification mechanism, enabling multiple tokens to be accepted per decoding step while avoiding the overhead of maintaining a separate draft model.

\item \textbf{\textit{ViSpec}}~\citep{kang2025vispecacceleratingvisionlanguagemodels}:  
ViSpec extends speculative decoding to vision-language models by introducing a lightweight vision adaptor in the drafting pathway. The adaptor compresses visual tokens into a compact representation, reducing redundant multimodal computation during drafting. This design enables efficient multimodal speculative decoding while preserving visual grounding through standard verification by the full model.

\item \textbf{\textit{MSD}}~\citep{lin2025speculativedecodingreimaginedmultimodal}:  
MSD focuses on training a strong multimodal draft model specifically designed for vision-language generation. It adopts a modality-aware architecture and a staged training strategy, where the model is first trained on text-only data and then adapted to multimodal inputs. This training procedure significantly improves draft token quality and allows the system to achieve substantial inference speedups under lossless speculative verification.

\end{itemize}

\subsection{Training-free Methods} Training-free speculative decoding methods accelerate inference without introducing additional trainable parameters or retraining the model. Instead, they exploit redundancy in the generation process by reusing previously generated text, matching substrings in the prompt, or constructing efficient data structures to retrieve candidate continuations. Because these methods do not require additional training, they are easy to integrate with existing models and systems.

\begin{itemize}[label=$\bullet$, leftmargin=*, itemsep=0pt, parsep=2pt]

\item \textbf{\textit{SAM Decoding}}~\citep{DBLP:conf/acl/HuWZZLCZ25}:  
SAM Decoding constructs a suffix automaton over previously generated text to enable efficient longest-suffix matching. When a match is found, the corresponding continuation can be retrieved as a draft sequence. These drafts are then verified using the standard speculative decoding mechanism, ensuring identical outputs while exploiting repetition during generation.


\item \textbf{\textit{Lookahead}}~\citep{DBLP:conf/kdd/ZhaoXLZG24}:  
Lookahead accelerates decoding by retrieving multi-token continuations from a trie constructed over the prompt and previously generated tokens. The retrieved candidates are organized into a tree and verified by the target model using a tree-based accept mechanism. This approach can accept multiple tokens per decoding step while maintaining correctness, with worst-case fallback to standard autoregressive decoding.

\item \textbf{\textit{Recycling}}~\citep{DBLP:conf/acl/LuoWZZZ0X25}:  
Recycling reuses previously discarded candidate tokens generated during earlier decoding steps. Instead of discarding these candidates, the method converts them into a reusable search space that can be leveraged to construct new draft trees. This approach exploits redundancy across decoding steps while remaining entirely training-free.

\item \textbf{\textit{Prompt Lookup Decoding (PLD)}}~\citep{saxena2023prompt}:  
Prompt Lookup Decoding replaces the draft model with an efficient n-gram lookup mechanism over the prompt and context. When a matching n-gram is found, the tokens following that span are proposed as draft candidates. The method is fully training-free and can support both greedy decoding and sampling, while preserving correctness through standard speculative verification.

\end{itemize}
\section{Experiments}

\subsection{Experiment Setup}
\textbf{Baselines.} Our main evaluations were conducted on Qwen2.5-VL-7B-Instruct and LLaVA-1.5-7B. We evaluate overall 10 speculative decoding methods covering three different types as discussed in Table~\ref{tab:speculative_methods_summary} on our six subtasks. We followed the default parameters as specified in their original implementations.

\noindent
\textbf{Evaluation Metrics:} Since the selected speculative decoding methods are all lossless, following the design from~\citep{xia2024unlocking}, our benchmark only contains two key efficiency metrics, including \textit{Mean Accepted Tokens (MAT)}, which measures the average number of tokens accepted per speculative decoding step, and the \textit{Walltime Speedup Ratio (Speed)}, which quantifies the inference efficiency gain relative to vanilla autoregressive decoding. 
To better understand the inner behavior of the algorithms, we also provide the latency experiment of baseline speculative decoding methods. 
All experiments are conducted on four NVIDIA A100 GPUs, we provide full evaluation details in the appendix.

\begin{table}[h]
  \centering
  \caption{Performance comparison of speculative decoding methods for Qwen2.5-VL-7B and LLaVA-1.5-7B. The highest and the second highest scores of methods are respectively highlighted in \blue{blue} and \red{red}.}
  \resizebox{\textwidth}{!}{
    \begin{tabular}{c|c|cccccccccccc|cc}
    \toprule[1.5pt]
    \multirow{2}[4]{*}{Model} & \multicolumn{1}{c|}{Subtask} 
    & \multicolumn{2}{c}{GQA} 
    & \multicolumn{2}{c}{TVQA} 
    & \multicolumn{2}{c}{IC} 
    & \multicolumn{2}{c}{CQA} 
    & \multicolumn{2}{c}{CR} 
    & \multicolumn{2}{c|}{MTC} 
    & \multicolumn{2}{c}{Overall} \\
    \cmidrule{2-16}          
    & Method      
    & MAT   & Speed 
    & MAT   & Speed 
    & MAT   & Speed 
    & MAT   & Speed 
    & MAT   & Speed 
    & MAT   & Speed 
    & MAT   & Speed \\
    \midrule

    \multirow{13}[0]{*}{\makecell{Qwen2.5-VL-7B}} 
    & \multicolumn{1}{c|}{AR Baseline} & - & 1$\times$ & - & 1$\times$ & - & 1$\times$ & - & 1$\times$ & - & 1$\times$ & - & 1$\times$ & - & 1$\times$ \\
    \cmidrule(lr){2-16} 

    \rowcolor[HTML]{F3F5F7}
    & \multicolumn{15}{c}{\textit{\textbf{Training-based Methods}}} \\
    \cmidrule(lr){2-16}
    & EAGLE-1 & 2.41 & 1.81$\times$ & 2.04 & 1.28$\times$ & 2.46 & 2.59$\times$ & 2.24 & 1.93$\times$ & 2.46 & 2.22$\times$ & 2.33 & 2.14$\times$ & \high{2.36} & \high{2.11$\times$} \\
    & EAGLE-2 & 1.79 & 1.73$\times$ & 1.52 & 1.17$\times$ & 1.70 & 2.32$\times$ & 1.87 & 1.95$\times$ & 1.86 & 2.56$\times$ & 1.78 & 1.92$\times$ & 1.78 & 2.02$\times$ \\
    & EAGLE-3 & 0.43 & 1.02$\times$ & 0.22 & 0.83$\times$ & 0.38 & 1.17$\times$ & 0.22 & 1.09$\times$ & 0.16 & 0.85$\times$ & 0.25 & 0.97$\times$ & 0.24 & 0.96$\times$ \\
    & Medusa  & 0.82 & 1.29$\times$ & 0.66 & 1.01$\times$ & 0.72 & 1.55$\times$ & 0.92 & 1.59$\times$ & 0.82 & 1.99$\times$ & 0.82 & 1.38$\times$ & 0.80 & 1.49$\times$ \\
    & MSD      & 2.32 & 2.27$\times$ & 2.37 & 1.80$\times$ & 2.35 & 2.03$\times$ & 2.67 & 1.96$\times$ & 2.42 & 4.06$\times$ & 2.51 & 2.78$\times$ & \best{2.57} & \best{2.58$\times$} \\
    & ViSpec  & 2.01 & 1.67$\times$ & 1.26 & 1.12$\times$ & 1.75 & 2.07$\times$ & 1.57 & 1.72$\times$ & 1.28 & 1.52$\times$ & 1.15 & 1.42$\times$ & 1.29 & 1.51$\times$ \\
    \cmidrule(lr){2-16} 

    \rowcolor[HTML]{F3F5F7}
    & \multicolumn{15}{c}{\textit{\textbf{Training-free Methods}}} \\
    \cmidrule(lr){2-16}

    & Lookahead & 0.20 & 1.06$\times$ & 0.07 & 0.77$\times$ & 0.11 & 1.08$\times$ & 0.43 & 1.34$\times$ & 0.65 & 1.47$\times$ & 0.30 & 0.95$\times$ & 0.33 & 1.07$\times$ \\
    & Recycling & 0.05 & 1.02$\times$ & 0.04 & 0.80$\times$ & 0.05 & 1.04$\times$ & 0.17 & 0.87$\times$ & 0.13 & 1.08$\times$ & 0.12 & 1.09$\times$ & 0.11 & 1.04$\times$ \\
    & PLD       & 0.03 & 1.01$\times$ & 0.00 & 0.78$\times$ & 0.00 & 1.00$\times$ & 0.29 & 1.04$\times$ & 0.26 & 1.16$\times$ & 0.20 & 1.06$\times$ & 0.17 & 1.05$\times$ \\
    & SAM        & 0.11 & 1.33$\times$ & 0.10 & 1.23$\times$ & 0.16 & 1.48$\times$ & 0.51 & 2.04$\times$ & 0.19 & 6.53$\times$ & 0.27 & 2.04$\times$ & 0.23 & 2.17$\times$ \\
    \midrule

    \multirow{13}[0]{*}{\makecell{LLaVA-1.5-7B}}
    & \multicolumn{1}{c|}{AR Baseline} & - & 1$\times$ & - & 1$\times$ & - & 1$\times$ & - & 1$\times$ & - & 1$\times$ & - & 1$\times$ & - & 1$\times$ \\
    \cmidrule(lr){2-16} 

    \rowcolor[HTML]{F3F5F7}
    & \multicolumn{15}{c}{\textit{\textbf{Training-based Methods}}} \\
    \cmidrule(lr){2-16}

    & EAGLE-1  & 0.89 & 1.17$\times$ & 0.68 & 1.12$\times$ & 0.73 & 1.19$\times$ & 0.81 & 1.21$\times$ & 0.76 & 1.19$\times$ & 0.77 & 1.25$\times$ & 0.76 & 1.22$\times$ \\
    & EAGLE-2  & 0.65 & 1.14$\times$ & 0.54 & 1.13$\times$ & 0.57 & 1.18$\times$ & 0.65 & 1.22$\times$ & 0.56 & 1.17$\times$ & 0.58 & 1.23$\times$ & 0.58 & 1.20$\times$ \\
    & EAGLE-3  & 0.06 & 0.68$\times$ & 0.07 & 0.71$\times$ & 0.10 & 0.73$\times$ & 0.06 & 0.72$\times$ & 0.01 & 0.68$\times$ & 0.03 & 0.69$\times$ & 0.04 & 0.70$\times$ \\
    & Medusa   & 1.12 & 1.56$\times$ & 0.77 & 1.65$\times$ & 0.84 & 1.32$\times$ & 0.95 & 1.88$\times$ & 1.07 & 1.66$\times$ & 1.52 & 1.06$\times$ & 1.30 & 1.26$\times$ \\
    & MSD      & 3.62 & 1.92$\times$ & 3.68 & 1.88$\times$ & 3.70 & 2.48$\times$ & 3.74 & 1.97$\times$ & 3.49 & 2.47$\times$ & 3.68 & 2.49$\times$ & \best{3.66} & \high{2.38$\times$} \\
    & ViSpec   & 2.52 & 1.83$\times$ & 2.34 & 2.00$\times$ & 3.44 & 2.83$\times$ & 3.01 & 2.40$\times$ & 3.00 & 2.59$\times$ & 3.01 & 2.68$\times$ & \high{3.00} & \best{2.58$\times$} \\
    \cmidrule(lr){2-16} 

    \rowcolor[HTML]{F3F5F7}
    & \multicolumn{15}{c}{\textit{\textbf{Training-free Methods}}} \\
    \cmidrule(lr){2-16}

    & Lookahead & 0.37 & 1.09$\times$ & 0.14 & 0.97$\times$ & 0.15 & 0.92$\times$ & 0.52 & 1.19$\times$ & 1.70 & 1.91$\times$ & 0.46 & 1.11$\times$ & 0.52 & 1.17$\times$ \\
    & Recycling & 0.31 & 1.60$\times$ & 0.19 & 1.87$\times$ & 1.12 & 1.05$\times$ & 0.94 & 0.72$\times$ & 0.62 & 1.29$\times$ & 1.00 & 2.37$\times$ & 0.88 & 1.54$\times$ \\
    & PLD       & 0.14 & 1.20$\times$ & 0.03 & 1.14$\times$ & 0.01 & 0.99$\times$ & 1.26 & 1.13$\times$ & 2.04 & 1.96$\times$ & 0.48 & 2.67$\times$ & 0.60 & 1.77$\times$ \\
    & SAM       & 0.31 & 1.09$\times$ & 0.11 & 0.93$\times$ & 0.07 & 0.87$\times$ & 0.42 & 1.14$\times$ & 1.08 & 1.63$\times$ & 0.34 & 1.11$\times$ & 0.38 & 1.13$\times$ \\
    \bottomrule[1.5pt]
    \end{tabular}}
  \label{tab:main_results}%
\end{table}

\subsection{Overall Comparison}
\label{subsec:discussion_on_results}
The overall comparison is illustrated in Table~\ref{tab:main_results}. We also provide the non-greedy overall comparison in the appendix. We have three key findings with one main takeaway as follows:

\noindent
\textbf{- Model-free speculative decoding methods show very limited benefits and sometimes even fail to achieve speedup.} We observe that model-free methods, including Lookahead, Recycling, PLD, and SAM, consistently achieve only marginal improvements over the AR baseline, and in some cases even lead to slowdowns. For example, Lookahead achieves only 1.07× speedup on Qwen2.5-VL-7B and 1.17× on LLaVA-1.5-7B overall. In several subtasks, the speedup even drops below 1×, indicating that the speculative mechanism introduces overhead without providing effective draft predictions. This phenomenon suggests that heuristic or history-based token prediction strategies are insufficient for the multimodal generation scenario. Unlike pure text generation, vision-language models involve complex cross-modal reasoning, where the token distribution is highly dependent on visual inputs. As a result, simple token reuse or lookahead strategies fail to generate reliable draft tokens, leading to low acceptance rates and limited acceleration.

\noindent
\textbf{- Training-based methods that ignore visual information also perform poorly.} Training a draft model alone does not guarantee effective acceleration in the multimodal setting. Methods such as EAGLE-1/2/3 and Medusa, which are originally designed for text-only LLMs, show limited improvements when directly applied to VLMs. For instance, on Qwen2.5-VL-7B, EAGLE-2 achieves only 2.02× speedup overall, while EAGLE-3 even drops below the baseline in some subtasks. Similar trends are observed for LLaVA-1.5-7B, where these methods provide only modest gains compared to specialized approaches. The main reason is that these methods only model the language decoding process while ignoring the visual-conditioned token distribution. In vision-language generation, the next-token prediction is strongly influenced by the encoded visual features. Draft models trained solely on textual signals therefore struggle to produce accurate predictions, resulting in low acceptance rates during verification.

\noindent
\textbf{- Even vision-aware training methods exhibit unstable performance across tasks.} Recent approaches that explicitly incorporate visual information during speculative decoding training, such as MSD and ViSpec, achieve higher average speedups compared to previous methods. However, their performance remains unstable across different subtasks and model architectures. For example, MSD achieves 2.58× speedup on Qwen2.5-VL-7B, but its performance varies significantly across tasks, ranging from 1.80× to over 4×. Similar instability can be observed for ViSpec, where the speedup fluctuates notably across subtasks. This instability indicates that current vision-aware speculative decoding techniques are still far from robust. The interaction between visual representations and language generation introduces additional complexity, and existing methods may struggle to generalize across different types of multi-modal reasoning tasks.

\begin{takeaway} All existing methods have performance degradation to handle vision tasks, especially when generating vision-informative tokens. 
\end{takeaway}

\subsection{Sensitivity Study} 
\begin{figure}
    \centering
    \includegraphics[width=1\linewidth]{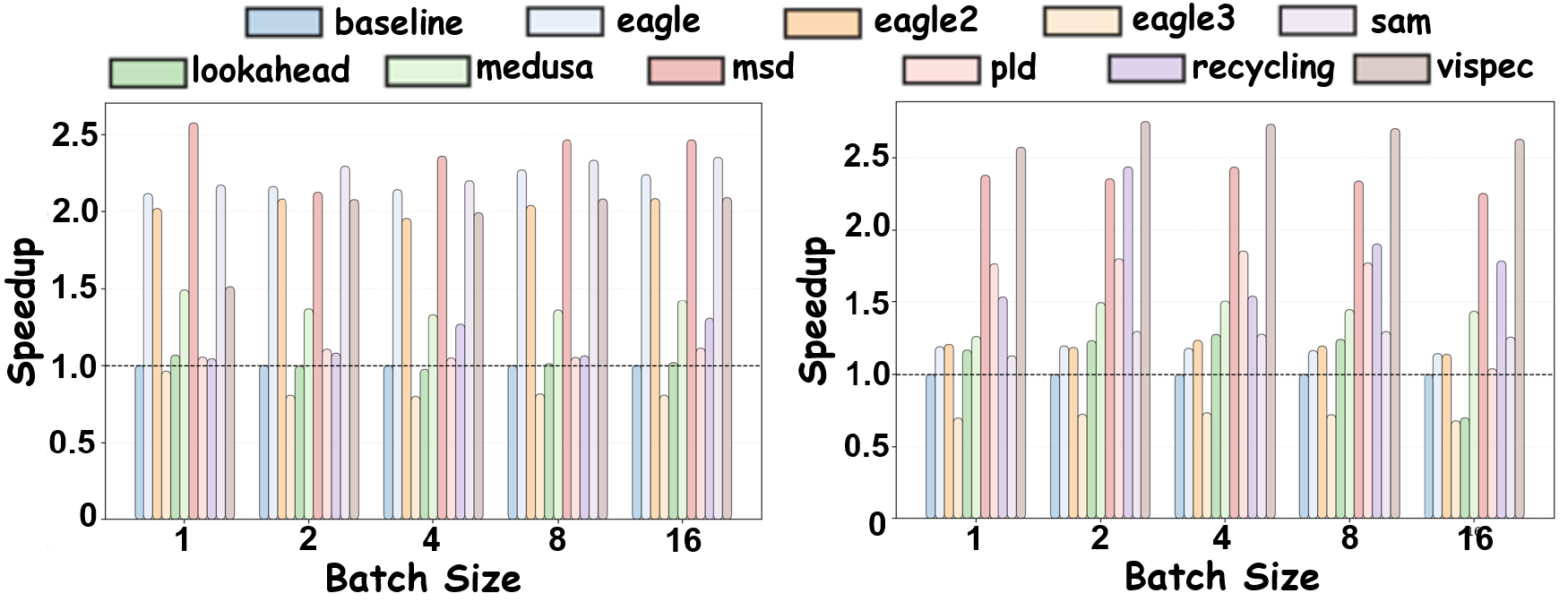}
    \caption{Speedup comparison of speculative decoding methods across different batch sizes. The speedup is measured relative to autoregressive decoding.}
    \label{fig:bz}
\end{figure}

We finally evaluate the sensitivity of different speculative decoding algorithms under various batch size. The results are provided in Figure~\ref{fig:bz}, and we have two observations with one key takeaway.

\noindent
\textbf{- Vision-aware methods consistently outperform other speculative decoding approaches.} Across all batch sizes and both models, vision-aware speculative decoding methods, such as MSD and ViSpec, consistently achieve the highest speedups. For example, MSD reaches around 2.3×–2.6× speedup on both Qwen2.5-VL-7B and LLaVA-1.5-7B across all batch sizes, while ViSpec also maintains competitive improvements. In contrast, speculative decoding methods originally designed for text-only LLMs show noticeably weaker performance. This result suggests that explicitly modeling the interaction between visual inputs and language generation significantly improves draft token quality, leading to higher acceptance rates and better overall acceleration.

\noindent
\textbf{- Non-vision-aware methods suffer significant performance degradation as batch size increases.} Another notable observation is that methods that do not incorporate visual awareness exhibit unstable or degraded performance as batch size increases. For instance, the speedup of methods such as EAGLE-2 and Lookahead decreases significantly when moving from small batch sizes to larger ones. This degradation likely arises because draft predictions become increasingly inaccurate in multimodal settings when visual information is ignored. As batch size increases, the system processes more heterogeneous multimodal inputs simultaneously, further amplifying the mismatch between the predicted tokens and the target model's outputs. Consequently, the acceptance rate drops, leading to reduced speculative decoding efficiency.

\begin{takeaway}
Vision awareness becomes increasingly important for speculative decoding in vision-language models, especially under larger batch sizes. As batching increases the diversity of multimodal inputs, methods that fail to incorporate visual information struggle to produce accurate draft tokens, resulting in diminished speedups. In contrast, vision-aware methods remain robust and maintain stable acceleration across different batch sizes.
\end{takeaway}

\subsection{Latency Analysis} 
\begin{figure}
    \centering
    \includegraphics[width=0.9\linewidth]{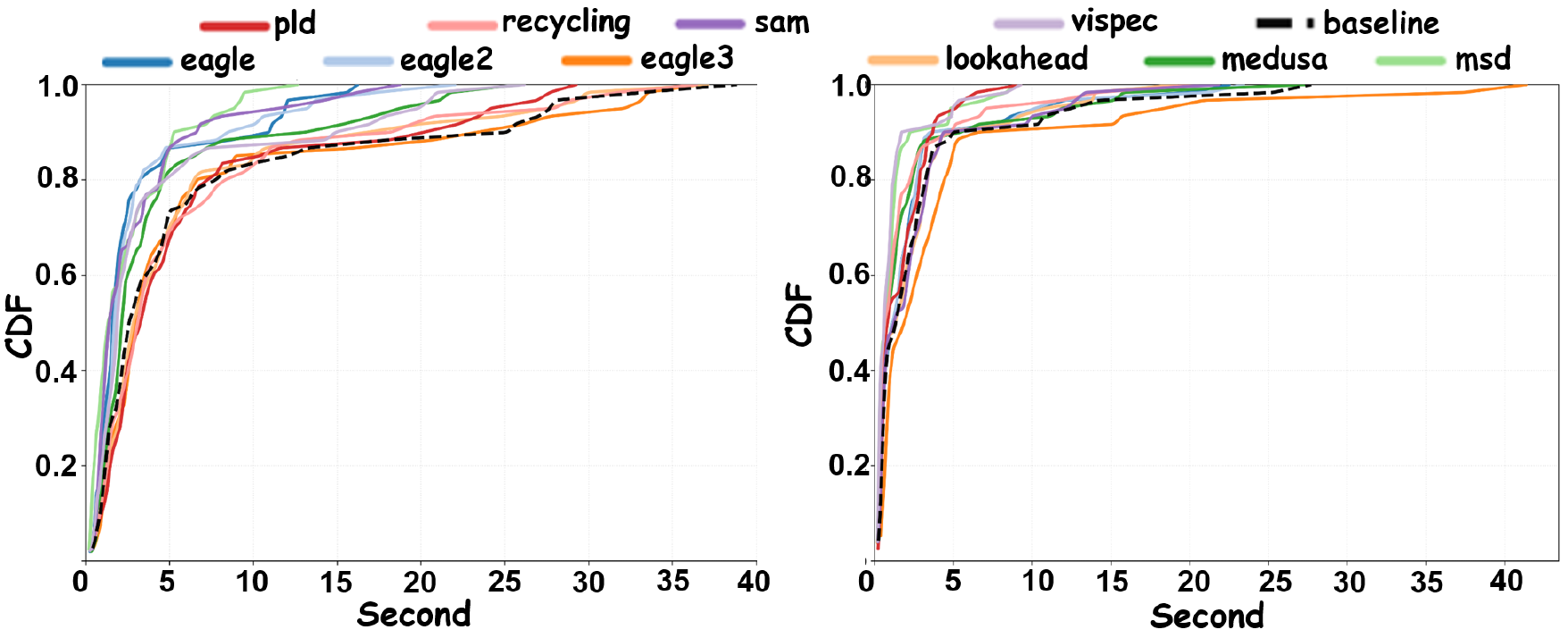}
    \caption{CDF of per-sample latency for different speculative decoding methods.}
    \label{fig:latency}
\end{figure}

We finally measure the latency of each requests processed by different speculative decoding algorithms. The results are shown in Figure~\ref{fig:latency}. We obtain two findings with one key takeaway.

\noindent
\textbf{- Vision-aware methods achieve consistently lower latency.}
Across both models, vision-aware speculative decoding methods, such as MSD and ViSpec, consistently shift the CDF curves to the left compared with other approaches. This indicates that these methods can complete a larger fraction (over 50\%) of requests within shorter wall-clock time, demonstrating their ability to generate higher-quality draft tokens under multimodal conditions.

\noindent
\textbf{- Higher throughput speedup does not always translate to better latency.}
Interestingly, some methods that achieve relatively high throughput speedups do not always exhibit the best latency performance in the CDF curves. This discrepancy suggests that while these methods may improve average decoding efficiency, they can still suffer from unstable behavior across samples, leading to longer latency for certain inputs.

\begin{takeaway}
These results highlight that speculative decoding methods should be evaluated from both throughput and latency perspectives. Maximizing throughput speedup alone is insufficient; achieving stable and consistent latency improvements is equally important for practical multimodal inference systems.
\end{takeaway}
\section{\method}
Based on the findings above, we introduce \method, a vision-aware speculative decoding framework that adaptively alternates between conventional decoding and speculative drafting based on the estimated visual relevance of the current token state. The pseudocode of \method is provided in Algorithm~\ref{algo:viskip}. Below we first describe the main process of \method, and then compare its performance with other baseline methods.

\begin{algorithm}[t]
\caption{Main Process of \method}
\label{algo:viskip}
\begin{algorithmic}[1]
\REQUIRE Full model $F$, draft model $D$, threshold $\tau$, draft length $K$
\FOR{each decoding step $t$}
    \STATE Compute cross-attention scores $A_t$
    \STATE Compute $S_t = \max_i A_t^{(i)}$
    \IF{$S_t \le \tau$}
        \STATE Draft up to $K$ tokens using $D$
        \STATE Verify with $F$ (standard speculative decoding)
    \ELSE
        \STATE Generate one token using $F$
    \ENDIF
\ENDFOR
\end{algorithmic}
\end{algorithm}

\subsection{Methods}
Let a vision-language model (VLM) consist of a visual encoder that produces visual tokens
$V = \{v_1, \dots, v_M\}$ and a language decoder that generates text tokens autoregressively.
At decoding step $t$, denote the decoder hidden state as $h_t$. \method consists of two key components, including estimating vision relevance and switching adaptive draft as follows.

\paragraph{Vision Relevance Estimation.}
We measure the dependency of the current decoding state on visual information
through cross-attention weights between $h_t$ and visual tokens.
Let the cross-attention distribution be:

\begin{figure*}[t]
\centering

\begin{subfigure}[t]{0.33\textwidth}
\centering
\includegraphics[width=\linewidth]{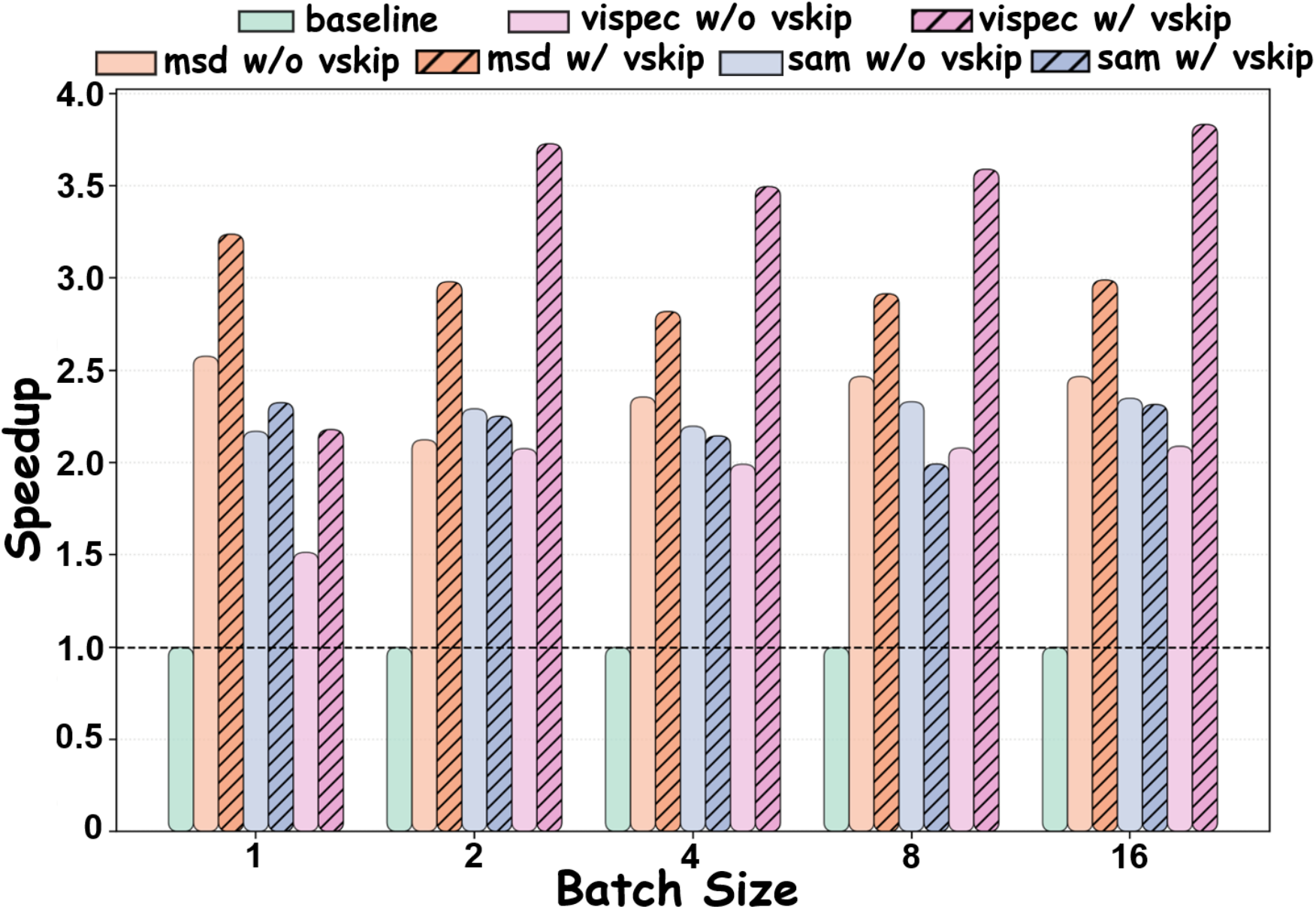}
\caption{}
\label{fig:viskip_a}
\end{subfigure}
\hfill
\begin{subfigure}[t]{0.32\textwidth}
\centering
\includegraphics[width=\linewidth]{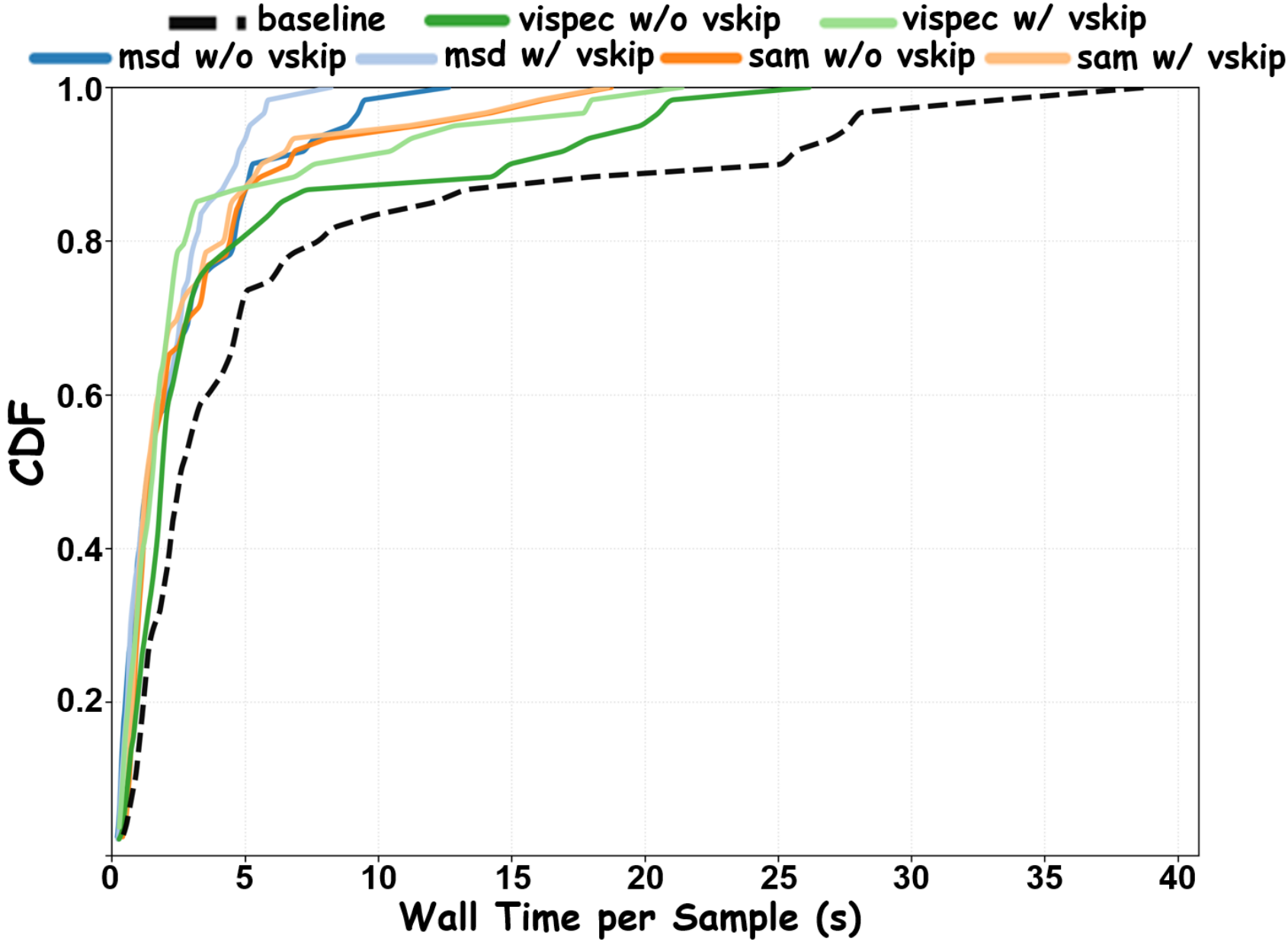}
\caption{}
\label{fig:viskip_b}
\end{subfigure}
\hfill
\begin{subfigure}[t]{0.32\textwidth}
\centering
\includegraphics[width=\linewidth]{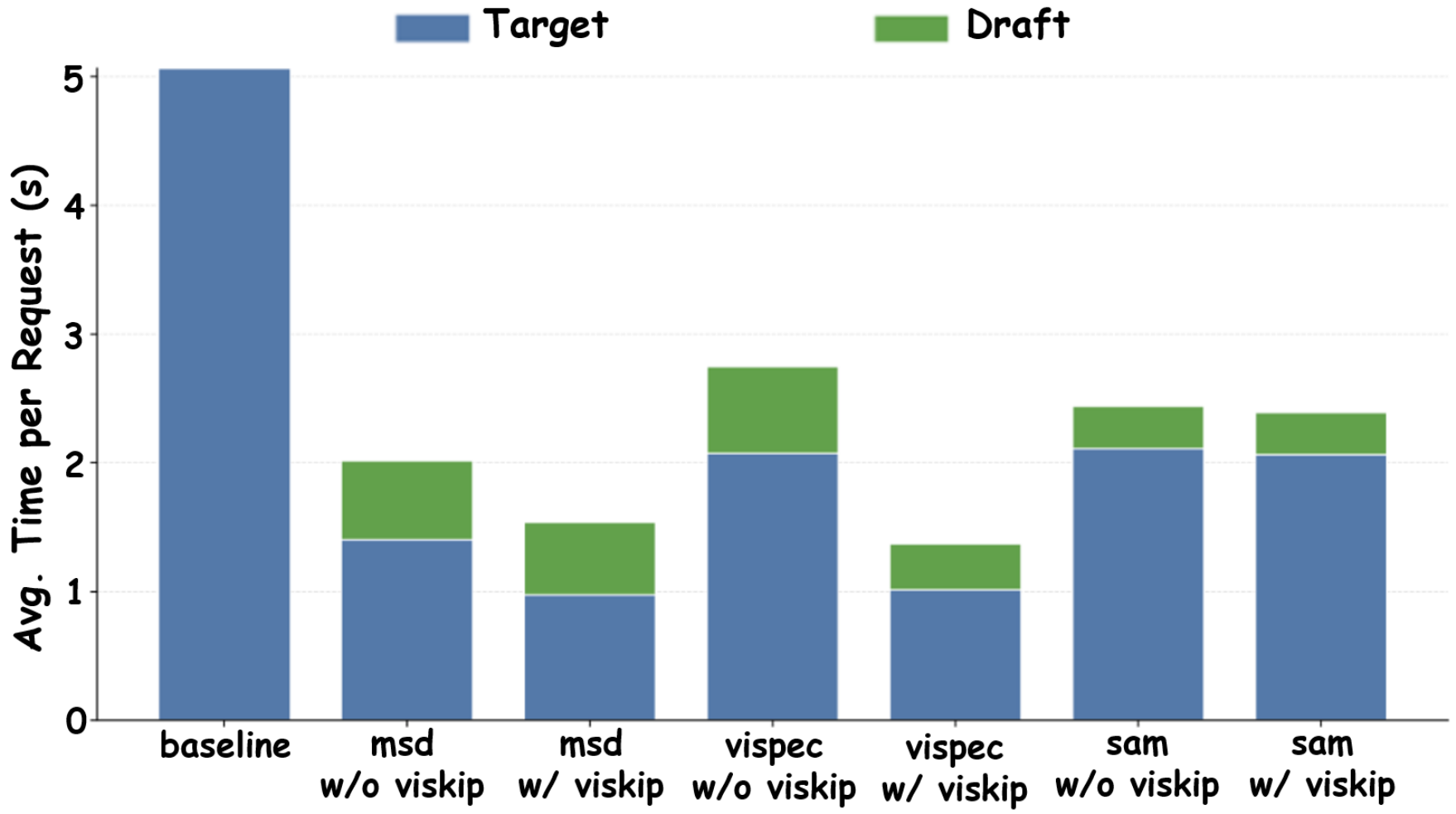}
\caption{}
\label{fig:viskip_results}
\end{subfigure}

\caption{Evaluation of ViSkip and baseline methods on Qwen2.5-VL-7B, including
(a) Speedup across different batch sizes. 
(b) CDF of wall time per sample. 
(c) Average latency breakdown of draft and target models.}

\label{fig:viskip_all}

\end{figure*}

\begin{equation}
A_t = \text{CrossAttn}(h_t, V),
\end{equation}

where $A_t^{(i)}$ denotes the attention weight assigned to visual token $v_i$. To capture the visual dependency at step $t$, we define a vision relevance score as:

\begin{equation}
S_t = \max_{i \in [1, M]} A_t^{(i)},
\end{equation}

Given threshold $\tau$, we determine whether the next token is vision-related by:

\begin{equation}
\mathbb{I}_{\text{vision}}(t) =
\begin{cases}
1 & \text{if } S_t > \tau, \\
0 & \text{otherwise}.
\end{cases}
\end{equation}

\paragraph{Adaptive Draft Switching.}
Let $F$ denote the full VLM and $D$ denote the draft model.
Given maximum draft length $K$, \method dynamically switch the decoding policy between normal and speculative decoding at step $t$ by:

\begin{equation}
\text{Decode}(t) =
\begin{cases}
\text{SpecDecode}(F, D, K) & \text{if } S_t \le \tau, \\
\text{Greedy}(F) & \text{if } S_t > \tau.
\end{cases}
\end{equation}

Specifically, if $S_t \le \tau$, \method performs standard speculative decoding: the draft model proposes up to $K$ tokens, which are verified by $F$. If $S_t > \tau$, the drafting is disabled and the full model generates a single token autoregressively.

\subsection{Evaluation Results}

To demonstrate the effectiveness of \method, we compare its performance with three state-of-the-art speculative decoding algorithms, including the training-based methods ViSpec and MSD and the training-free method SAM Decoding. Due to the page limit, we measure the impact of $\tau$ on the performance of \method in the Appendix. Since \method is a plug-and-play technique, we evaluate these algorithms both with and without \method. The results are shown in Fig.~\ref{fig:viskip_results}. As illustrated in Fig.~\ref{fig:viskip_results}(a), integrating \method consistently improves the speedup of all three methods across different batch sizes. Fig.~\ref{fig:viskip_results}(b) further shows that the latency CDF curves shift to the left after applying \method, indicating faster completion for most samples. The latency breakdown in Fig.~\ref{fig:viskip_results}(c) reveals that \method significantly reduces the computation time of the target model, leading to the overall speedup improvement.


\section{Conclusion}
We present \dataset, the first benchmark for systematically evaluating speculative decoding in vision-language models. Our study reveals that existing methods often degrade in multimodal settings, vision awareness becomes increasingly important at larger batch sizes, and throughput speedup alone is insufficient to evaluate practical efficiency. Based on these findings, we propose \method, a plug-and-play speculative decoding method that dynamically adapts speculation to vision tokens and consistently improves existing approaches. We hope \dataset and the insights from this study can guide future research on efficient multimodal generation. 

%
%
\bibliographystyle{plainnat}
\bibliography{main}

\appendix
\clearpage

\section{Appendix}

\subsection{Performance Comparison with Different Temperature}

\begin{table}[h]
  \centering
  \caption{Performance comparison of speculative decoding methods for Qwen2.5-VL-7B and LLaVA-1.5-7B at temperature $t=0.6$. The highest and the second highest overall scores are respectively highlighted in \blue{blue} and \red{red}.}
  \resizebox{\textwidth}{!}{
    \begin{tabular}{c|c|cccccccccccc|cc}
    \toprule[1.5pt]
    \multirow{2}[4]{*}{Model} & \multicolumn{1}{c|}{Subtask}
    & \multicolumn{2}{c}{GQA}
    & \multicolumn{2}{c}{TVQA}
    & \multicolumn{2}{c}{IC}
    & \multicolumn{2}{c}{CQA}
    & \multicolumn{2}{c}{CR}
    & \multicolumn{2}{c|}{MTC}
    & \multicolumn{2}{c}{Overall} \\
    \cmidrule{2-16}
    & Method
    & MAT   & Speed
    & MAT   & Speed
    & MAT   & Speed
    & MAT   & Speed
    & MAT   & Speed
    & MAT   & Speed
    & MAT   & Speed \\
    \midrule

    \multirow{11}[0]{*}{\makecell{Qwen2.5-VL-7B}}
    & \multicolumn{1}{c|}{AR Baseline} & - & 1$\times$ & - & 1$\times$ & - & 1$\times$ & - & 1$\times$ & - & 1$\times$ & - & 1$\times$ & - & 1$\times$ \\
    \cmidrule(lr){2-16}

    \rowcolor[HTML]{F3F5F7}
    & \multicolumn{15}{c}{\textit{\textbf{Training-based Methods}}} \\
    \cmidrule(lr){2-16}
    & EAGLE-1 & 2.29 & 1.56$\times$ & 1.80 & 1.09$\times$ & 2.22 & 2.11$\times$ & 2.30 & 2.01$\times$ & 2.26 & 2.08$\times$ & 1.99 & 1.65$\times$ & \best{2.09} & 1.76$\times$ \\
    & EAGLE-2 & 1.59 & 1.42$\times$ & 1.39 & 1.07$\times$ & 1.59 & 1.66$\times$ & 1.77 & 1.75$\times$ & 1.70 & 1.76$\times$ & 1.67 & 1.65$\times$ & 1.66 & 1.63$\times$ \\
    & EAGLE-3 & 0.03 & 0.71$\times$ & 0.01 & 0.57$\times$ & 0.08 & 1.11$\times$ & 0.03 & 0.98$\times$ & 0.13 & 0.72$\times$ & 0.03 & 1.41$\times$ & 0.07 & 1.02$\times$ \\
    & Medusa  & 0.84 & 1.31$\times$ & 0.71 & 1.00$\times$ & 0.69 & 1.25$\times$ & 0.87 & 1.30$\times$ & 0.78 & 1.39$\times$ & 0.78 & 1.35$\times$ & 0.77 & 1.32$\times$ \\
    & MSD     & 1.86 & 1.30$\times$ & 1.86 & 1.57$\times$ & 1.86 & 1.45$\times$ & 1.94 & 1.64$\times$ & 1.84 & 1.89$\times$ & 1.85 & 2.05$\times$ & 1.87 & 1.82$\times$ \\
    & ViSpec  & 2.32 & 1.73$\times$ & 2.03 & 1.15$\times$ & 2.12 & 1.95$\times$ & 2.16 & 2.15$\times$ & 1.98 & 2.07$\times$ & 2.10 & 1.80$\times$ & \high{2.08} & \high{1.84$\times$} \\
    \cmidrule(lr){2-16}

    \rowcolor[HTML]{F3F5F7}
    & \multicolumn{15}{c}{\textit{\textbf{Training-free Methods}}} \\
    \cmidrule(lr){2-16}
    & Lookahead & 0.21 & 1.06$\times$ & 0.06 & 0.77$\times$ & 0.09 & 1.00$\times$ & 0.42 & 1.47$\times$ & 0.56 & 1.50$\times$ & 0.35 & 0.99$\times$ & 0.33 & 1.09$\times$ \\
    & Recycling & 0.05 & 0.93$\times$ & 0.04 & 0.84$\times$ & 0.06 & 1.02$\times$ & 0.32 & 0.72$\times$ & 0.12 & 1.62$\times$ & 0.11 & 1.05$\times$ & 0.12 & 1.08$\times$ \\
    & PLD       & 0.03 & 0.85$\times$ & 0.00 & 0.72$\times$ & 0.00 & 0.97$\times$ & 0.27 & 1.03$\times$ & 0.20 & 1.24$\times$ & 0.24 & 0.99$\times$ & 0.18 & 1.02$\times$ \\
    & SAM       & 0.12 & 1.12$\times$ & 0.18 & 0.89$\times$ & 0.13 & 1.22$\times$ & 0.38 & 1.49$\times$ & 0.34 & 3.56$\times$ & 0.28 & 2.42$\times$ & 0.24 & \best{1.99$\times$} \\
    \midrule

    \multirow{11}[0]{*}{\makecell{LLaVA-1.5-7B}}
    & \multicolumn{1}{c|}{AR Baseline} & - & 1$\times$ & - & 1$\times$ & - & 1$\times$ & - & 1$\times$ & - & 1$\times$ & - & 1$\times$ & - & 1$\times$ \\
    \cmidrule(lr){2-16}

    \rowcolor[HTML]{F3F5F7}
    & \multicolumn{15}{c}{\textit{\textbf{Training-based Methods}}} \\
    \cmidrule(lr){2-16}
    & EAGLE-1 & 0.11 & 0.86$\times$ & 0.06 & 0.73$\times$ & 0.07 & 0.78$\times$ & 0.05 & 0.65$\times$ & 0.05 & 1.02$\times$ & 0.06 & 0.65$\times$ & 0.06 & 0.74$\times$ \\
    & EAGLE-2 & 0.06 & 0.81$\times$ & 0.03 & 0.78$\times$ & 0.03 & 0.70$\times$ & 0.06 & 0.88$\times$ & 0.05 & 1.83$\times$ & 0.05 & 0.66$\times$ & 0.05 & 0.81$\times$ \\
    & EAGLE-3 & 0.02 & 0.79$\times$ & 0.06 & 0.82$\times$ & 0.10 & 0.74$\times$ & 0.10 & 0.87$\times$ & 0.01 & 1.67$\times$ & 0.05 & 0.85$\times$ & 0.06 & 0.91$\times$ \\
    & Medusa  & 0.94 & 1.63$\times$ & 0.93 & 1.48$\times$ & 0.92 & 1.30$\times$ & 1.09 & 1.57$\times$ & 0.93 & 3.33$\times$ & 1.15 & 1.85$\times$ & 1.03 & \high{1.80$\times$} \\
    & MSD     & 2.77 & 1.74$\times$ & 2.95 & 1.56$\times$ & 2.98 & 1.98$\times$ & 3.39 & 1.99$\times$ & 2.87 & 2.95$\times$ & 2.99 & 1.54$\times$ & \best{2.99} & \best{1.85$\times$} \\
    & ViSpec  & 1.60 & 0.82$\times$ & 1.39 & 2.11$\times$ & 1.28 & 1.74$\times$ & 1.59 & 1.15$\times$ & 1.75 & 4.55$\times$ & 1.41 & 1.43$\times$ & 1.45 & 1.68$\times$ \\
    \cmidrule(lr){2-16}

    \rowcolor[HTML]{F3F5F7}
    & \multicolumn{15}{c}{\textit{\textbf{Training-free Methods}}} \\
    \cmidrule(lr){2-16}
    & Lookahead & 0.41 & 1.03$\times$ & 0.14 & 1.06$\times$ & 0.14 & 0.95$\times$ & 0.53 & 1.45$\times$ & 1.70 & 1.63$\times$ & 0.46 & 0.70$\times$ & 0.52 & 0.94$\times$ \\
    & Recycling & 0.32 & 1.26$\times$ & 0.95 & 1.67$\times$ & 0.52 & 2.36$\times$ & 1.08 & 0.69$\times$ & 2.38 & 2.54$\times$ & 2.22 & 1.33$\times$ & \high{1.60} & 1.46$\times$ \\
    & PLD       & 0.10 & 1.11$\times$ & 0.02 & 0.81$\times$ & 0.02 & 0.97$\times$ & 1.18 & 1.36$\times$ & 0.67 & 2.42$\times$ & 0.28 & 1.09$\times$ & 0.30 & 1.19$\times$ \\
    & SAM       & 0.32 & 0.93$\times$ & 0.08 & 0.89$\times$ & 0.08 & 0.89$\times$ & 0.42 & 1.28$\times$ & 0.49 & 1.03$\times$ & 0.36 & 0.71$\times$ & 0.32 & 0.85$\times$ \\
    \bottomrule[1.5pt]
    \end{tabular}}
  \label{tab:main_results_t06}%
\end{table}

\noindent
Compared with greedy decoding ($t=0$), sampling at $t=0.6$ generally makes speculative decoding less stable, with a clearer degradation for training-based methods. On Qwen2.5-VL-7B, most training-based methods show lower overall MAT and speed at $t=0.6$, such as MSD (\(2.57/2.58\times \rightarrow 1.87/1.82\times\)) and EAGLE-1 (\(2.36/2.11\times \rightarrow 2.09/1.76\times\)), indicating that higher temperature weakens draft-target agreement. On LLaVA-1.5-7B, the same trend is even stronger: EAGLE-1/2 collapse substantially, MSD drops from \(3.66/2.38\times\) to \(2.99/1.85\times\), and ViSpec decreases from \(3.00/2.58\times\) to \(1.45/1.68\times\). In contrast, training-free methods remain relatively conservative across temperatures. These results suggest that increasing temperature mainly harms the reliability of training-based speculative drafts, whereas training-free methods have lower ceilings but tend to be less sensitive to sampling noise.

\subsection{Implementation Details}
Tables~\ref{tab:main_exp_shared} and \ref{tab:main_exp_method_params} summarize the settings used by the main MMSpec experiments reported in the paper. Unless otherwise stated, the method-specific hyperparameters are identical for $t=0$ and $t=0.6$.

\begin{table}[t]
  \centering
  \caption{Shared settings for the main \dataset experiments.}
  \label{tab:main_exp_shared}
  \begin{tabular}{p{0.28\linewidth} p{0.62\linewidth}}
    \toprule
    Item & Setting \\
    \midrule
    Evaluated models & \texttt{Qwen/Qwen2.5-VL-7B-Instruct} \texttt{llava-hf/llava-1.5-7b-hf} \\
    Batch size & 1 \\
    Max new tokens & 1024 \\
    Decoding temperatures & $t \in \{0.0, 0.6\}$ \\
    \bottomrule
  \end{tabular}
\end{table}

\begin{table*}[t]
  \centering
  \caption{Method-specific parameters used in the main experiments. ``TB'' and ``TF'' denote training-based and training-free methods, respectively.}
  \label{tab:main_exp_method_params}
  \small
  \setlength{\tabcolsep}{4pt}
  \begin{tabular}{p{0.12\linewidth} p{0.08\linewidth} p{0.68\linewidth}}
    \toprule
    Method & Type & Hyperparameters used in main runs \\
    \midrule
    EAGLE-1 & TB & \texttt{depth=3}, \texttt{top\_k=8}, \texttt{total\_token=30} \\
    EAGLE-2 & TB & \texttt{depth=3}, \texttt{top\_k=8}, \texttt{total\_token=30}, \texttt{threshold=0.3} \\
    EAGLE-3 & TB & \texttt{depth=3}, \texttt{top\_k=8}, \texttt{total\_token=30} \\
    Medusa & TB & \texttt{depth=3}, \texttt{top\_k=8}, \texttt{total\_token=30} \\
    MSD & TB & \texttt{depth=5}, \texttt{top\_k=10} \\
    ViSpec & TB & \texttt{depth=3}, \texttt{top\_k=8}, \texttt{total\_token=30}, \texttt{num\_q=2} \\
    Lookahead & TF & \texttt{decoding\_length=64}, \texttt{branch\_length=12} \\
    Recycling & TF & \texttt{matrix\_top\_k=8}, \texttt{draft\_len=10} \\
    PLD & TF & \texttt{ngram=4}, \texttt{n\_pred=10} \\
    SAM & TF & \texttt{total\_token=30}, \texttt{depth=3}, \texttt{top\_k=8}, \texttt{threshold=1.0} \\
    \bottomrule
  \end{tabular}
\end{table*}




\subsection{Quantitative Analysis of Acceptance on Vision-Grounded Steps}
We run MSD, ViSpec, and SAM while logging, for every decoding step, both the speculative acceptance length and the visual-attention ratio computed by the same attention-scoring rule used in ViSkip. We then define a \emph{high-visual} step as one whose visual-attention ratio is at least $0.35$, which matches the ViSkip threshold used in our probe runs.

\begin{table*}[t]
  \centering
  \caption{Step-level acceptance statistics on Qwen2.5-VL-7B from the visual-attention probe runs on \dataset testmini set. ``Measured steps'' are steps with a valid visual-attention ratio.}
  \label{tab:qwen_visual_probe_quant}
  \small
  \setlength{\tabcolsep}{3.5pt}
  \renewcommand{\arraystretch}{1.1}
  \resizebox{\linewidth}{!}{
  \begin{tabular}{l c c c c c}
    \toprule
    Method 
    & Measured steps 
    & High-visual steps 
    & \makecell[c]{Avg.\ accept\\(high-visual)} 
    & \makecell[c]{Avg.\ accept\\(low-visual)}  \\
    \midrule
    MSD    & 2170 & 332 (15.3\%) & 2.12 & 2.27  \\
    ViSpec & 1874 & 170 (9.1\%)  & 2.12 & 2.39  \\
    SAM    & 2469 & 176 (7.1\%)  & 0.14 & 0.44 \\
    \bottomrule
  \end{tabular}}
\end{table*}

Table~\ref{tab:qwen_visual_probe_quant} shows a clear quantitative trend. For all three methods, steps with stronger attention to image tokens are associated with worse speculative acceptance. For MSD, the effect is moderate but consistent: the average acceptance length drops from 2.27 on low-visual steps to 2.12 on high-visual steps. ViSpec shows the same pattern: its average acceptance length drops from 2.39 to 2.12. For SAM, the effect is much stronger: the average acceptance length drops from 0.44 to 0.14. In other words, once the next token is more visually grounded, SAM is very likely to accept nothing from the draft model at all.

Overall, the quantitative picture is consistent with the motivation of ViSkip: speculative decoding is least reliable exactly when the next decoding step depends more strongly on visual evidence.

\paragraph{Takeaway.}
When VLM attends more strongly to image tokens, speculative acceptance becomes worse, often collapsing to zero for visually grounded words or phrases. This provides direct empirical support for the core idea behind ViSkip.

\end{document}